\pdfoutput=1

\documentclass{bmvc2k}

\newcommand\blfootnote[1]{%
  \begingroup
  \renewcommand\thefootnote{}\footnote{#1}%
  \addtocounter{footnote}{-1}%
  \endgroup
}

\addauthor{Andrew Zhai*}{andrew@pinterest.com}{1}
\addauthor{Hao-Yu Wu*}{rexwu@pinterest.com}{1}

\addinstitution{
 Pinterest, Inc.\\
 San Francisco, US
}

\runninghead{Zhai, Wu}{Classification is a Strong Baseline for Deep Metric Learning}

\begin{document}

\title{Classification is a Strong Baseline for Deep Metric Learning}

\maketitle
\blfootnote{*These two authors contributed equally to this work}

\begin{abstract}
Deep metric learning aims to learn a function mapping image pixels to embedding feature vectors that model the similarity between images. Two major applications of metric learning are content-based image retrieval and face verification. For the retrieval tasks, the majority of current state-of-the-art (SOTA) approaches are triplet-based non-parametric training. For the face verification tasks, however, recent SOTA approaches have adopted classification-based parametric training. In this paper, we look into the effectiveness of classification based approaches on image retrieval datasets. We evaluate on several standard retrieval datasets such as CAR-196, CUB-200-2011, Stanford Online Product, and In-Shop datasets for image retrieval and clustering, and establish that our classification-based approach is competitive across different feature dimensions and base feature networks. We further provide insights into the performance effects of subsampling classes for scalable classification-based training, and the effects of binarization, enabling efficient storage and computation for practical applications.
\end{abstract}

\section{Introduction}
Metric learning, also known as learning image embeddings, is a core problem of a variety of applications including face recognition~\cite{facenet, Liu2017SphereFaceDH,Wang2018AdditiveMS,Wang2018CosFaceLM}, fine-grained retrieval~\cite{songCVPR16,samplingmatters,npairNIPS2016}, clustering~\cite{googleswirl}, and visual search~\cite{visualdiscoverypinterest,jing15,visualsearchalibaba,visualsearchbing}. The goal of metric learning is that the learned embedding generalize well to novel instances during test time, an open-set setup in machine learning. This goal aligns well with practical applications in which the deployed machine learning system is required to handle large amount of unseen data. 

Standard deep neural network metric learning methods train image embeddings through the local relationships between images in the form of \textit{pairs}~\cite{siamese,bell15productnet} or \textit{triplets}~\cite{triplet,facenet}. A core challenge with metric learning is sampling informative samples for training. As described in~\cite{facenet,defensetriplet,samplingmatters}, negatives that are too hard can destabilize training, while negatives that are too easy result in triplets that have near zero loss leading to slow convergence.

Recent methods such as~\cite{songCVPR16,npairNIPS2016,defensetriplet,samplingmatters} focus on addressing this sampling problem, many of which utilize the relationships of all images within the \textit{batch} to form informative triplets. These methods typically require a large batch size so that informative triplets can be selected within the batch. In practice, batch size is constrained by the hardware memory size. Therefore, as the dataset size grows, one still faces the challenge of the diminishing probability of a randomly sampled batch containing any informative triplet.

To alleviate the challenge of sampling informative triplets, another school of deep metric learning approaches propose to use classification-based training~\cite{nofusslearning,Wang2018CosFaceLM,Wang2018AdditiveMS,Liu2017SphereFaceDH,Xuan2018DeepRE}. In contrast to triplet-based approaches, the classification-based approach can be viewed as approximating each class using a proxy~\cite{nofusslearning}, and uses all proxies to provide global context for each training iteration. Though classification-based metric learning simplify training by removing sampling, they have scalability limitations as in the \textit{extreme classification}~\cite{songCVPR16} problem, where it is considered impractical to extend to much larger number of classes during training.

While generic deep metric learning approaches have been pushing SOTA through better negative sampling and ensembles of triplet-based approaches, face verification has in parallel seen SOTA results through classification-based loss functions. We set out to investigate if classification-based training can perform similarly well for general image retrieval tasks. 

Our major contributions are as follows: 1) we establish that classification is a strong baseline for deep metric learning across different datasets, base feature networks and embedding dimensions, 2) we provide insights into the performance effects of binarization and subsampling classes for scalable \textit{extreme classification}-based training 3) we propose a classification-based approach to learn high-dimensional binary embeddings that achieve state-of-the-art retrieval performance with the same memory footprint as 64 dimensional float embeddings across the suite of retrieval tasks.
\begin{figure*}
\centering
  \includegraphics[width=0.75\linewidth]{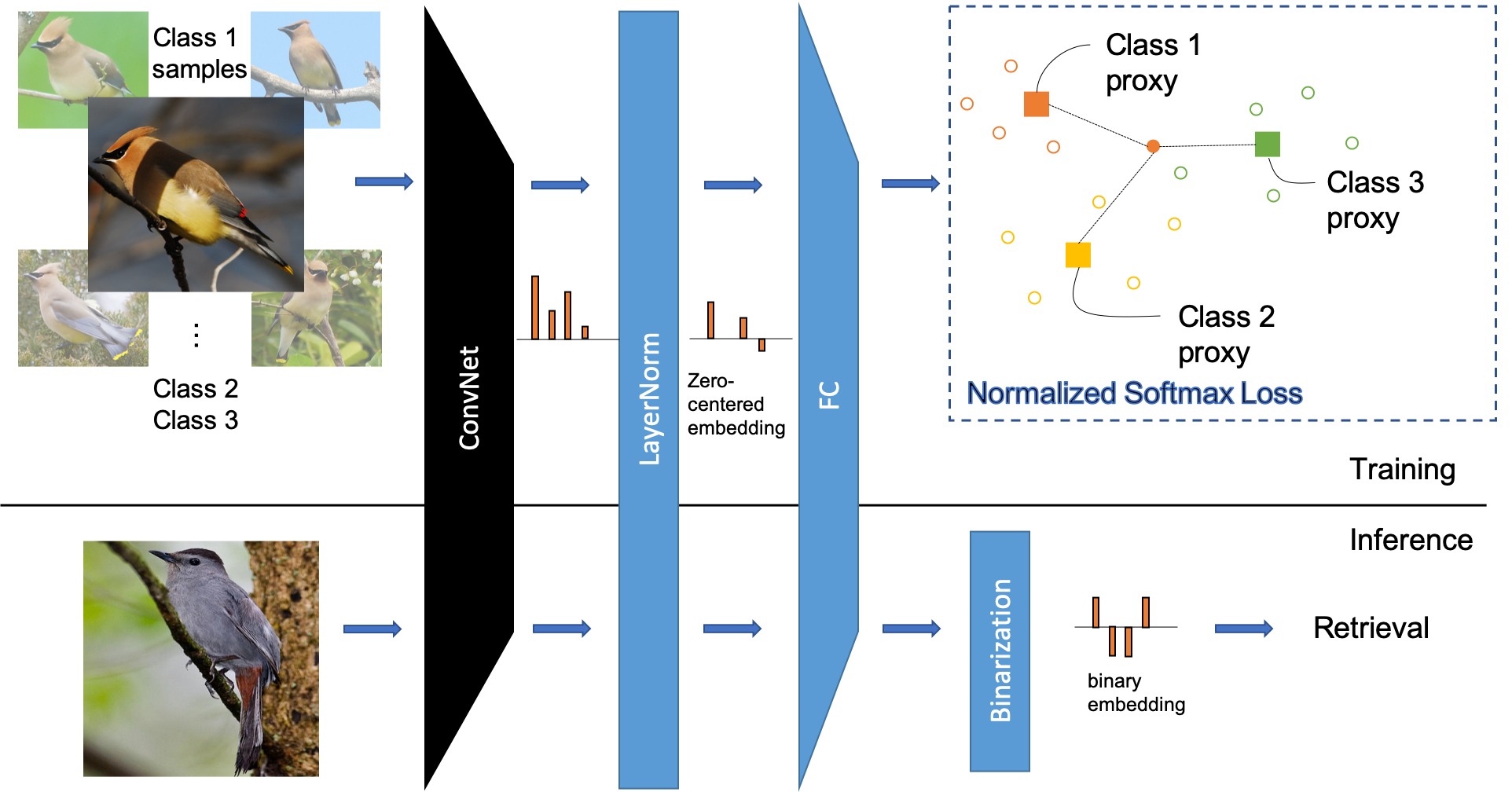}
  \caption{Architecture overview for training high dimensional binary embedding}
  \label{fig:overview}
\end{figure*}
\section{Related Works}
\label{subsec:related_metric}
\textbf{Metric Learning Losses}
Metric learning approaches aim to learn a good embedding space such that the similarity between samples are preserved as distance between embedding vectors of the samples. The metric learning losses, such as contrastive loss\cite{siamese} and triplet loss\cite{triplet}, are formulated to minimize intra-class distances and maximize inter-class distances. Recent approaches in metric learning design the loss function to consider the relationships of all the samples within the training batch\cite{songCVPR16,npairNIPS2016,angular,samplingmatters,UstinovaNIPS16}, and achieve state-of-the-art performance in image retrieval datasets\cite{cub200,cars196,songCVPR16}. 

\textbf{Training Sampling} 
Sampling informative training samples plays an important role in metric learning as also suggested in~\cite{facenet,samplingmatters}. Semi-hard sampling in conjunction with triplet-loss~\cite{facenet} has been widely adopted for many tasks. Distanced-weighted sampling~\cite{samplingmatters} suggests that with a balanced mix of difficulties during training, the image retrieval performance can be further improved. Hierarchical Triplet Loss~\cite{htl} proposed that by merging similar classes dynamically during training into a hierarchical tree, more informative samples can be drawn from such a structure and the loss also provides global context for training.

\textbf{Classification Losses for Metric Learning}
Classification losses are widely adopted in face verification applications~\cite{Liu2017SphereFaceDH,Wang2018AdditiveMS,Wang2018CosFaceLM} and achieve state-of-the-art performance. The theoretical link between classification and metric learning is shown in~\cite{nofusslearning}, and image retrieval tasks have some success adopting classification loss~\cite{nofusslearning,Xuan2018DeepRE}. Though classification-based training alleviates the need for carefully choosing sampling method, the parametric nature may cause issue in fine-grained open-set recognition setting~\cite{scalablenca}.

\textbf{Ensembling}
Ensembling embeddings has been the most recent focus to further improve image retrieval performance. The ensembled embeddings are trained via boosting~\cite{abier}, attending diverse spatial locations~\cite{attention_metric}, or partitioning the training classes~\cite{Xuan2018DeepRE}. However, such ensembled embeddings trade off image retrieval performance with higher dimensions. They also introduce additional hyperparameters into the system.


\section{Method} \label{sec:model}
We study recent model architectures and triplet-based and classification losses in Section~\ref{sec:experiments}. Beyond standard classification training~\cite{kaiming16}, we describe  below techniques we used to achieve SOTA on the retrieval tasks including L2 normalization of embedding to optimize for cosine similarity, Layer Normalization of final pooled feature layer, and class balanced sampling of minibatch. An overview of our approach is shown in Figure~\ref{fig:overview}.

\subsection{Normalized Softmax Loss}
When training the classification network for metric learning, we remove the bias term in the last linear layer and add an L2 normalization module to the inputs and weights before softmax loss to optimize for cosine similarity. Temperature scaling is used for exaggerating the difference among classes and boosting the gradients as also used in~\cite{scalablenca, Wang2018CosFaceLM,Liu2017SphereFaceDH}. We follow the same derivation and notations as in ~\cite{nofusslearning}. For embedding $x$ of input image with class label $y$, the loss with temperature $\sigma$ can be expressed with the weight of class y $p_y$ among all possible classes set $Z$:
\begin{equation}
\label{eq:proxy_cls_loss}
    L_{\text{norm}}
    = -\log\left(\frac{\exp(x^Tp_y / \sigma))}{\sum_{z \in Z} \exp(x^Tp_z/ \sigma))}\right)
\end{equation}
Normalized softmax loss fits into the proxy paradigm when we view the class weight as proxy and choose the distance metric as cosine distance function. A more general form of classification loss, Large Margin Cosine Loss (LMCL), has been proposed for face verification~\cite{Wang2018CosFaceLM} with an additional margin hyperparameter. We also evaluate the effect of LMCL in conjunction with our proposed techniques on image retrieval tasks in Section~\ref{sec:exp_soa}.
 
\subsection{Layer Normalization}
The layer normalization without affine parameters\cite{layernorm} is added immediately after the final pooling layer of the feature model (e.g. GoogleNet~\cite{googlenet2014}'s pool5 layer) to normalize the feature dimension of our embeddings to have a distribution of values centered at zero. This allows us to easily binarize embeddings via thresholding at zero. We also show empirically through ablation experiments in Section~\ref{sec:layer_norm_exp} that layer normalization helps the network better initialize new parameters and reach better optima.

\subsection{Class Balanced Sampling}
\label{subsec:sampling}
Based on the proxy explanation of using classification loss for metric learning, the loss is bounded by the worst approximated examples within the class~\cite{nofusslearning}. A simple way to alleviate this is by including multiple examples per class when constructing the training mini batch. For each training batch, we first sample $C$ classes, and sample $S$ samples within each class. The effect of class balanced sampling is studied through ablation experiment in Section ~\ref{sec:class_balance_exp} and is a common approach to retrieval tasks~\cite{samplingmatters, npairNIPS2016}.


\section{Experiments}
\label{sec:experiments}

We follow the same evaluation protocol commonly used in image retrieval tasks with the standard train/test split on four datasets: CARS-196~\cite{cars196}, CUB-200-2011~\cite{cub200}, Stanford Online Products (SOP)~\cite{songCVPR16}, and In-shop Clothes Retrieval~\cite{liu2016deepfashion}. We compare our method using Recall@K to measure retrieval quality. To compute Recall@K, we use cosine similarity to retrieve the top K images from the test set, excluding the query image itself.

We first investigate how our embeddings trained with normalized softmax loss compare against embeddings trained with existing metric learning losses using the same featurizer and embedding dimension. We then study in detail how the dimensionality of our embeddings (Section~\ref{sec:exp_dimension}) affects its performance and the relative performance between float and binary embeddings. Ablation studies on different design choices of our approach on CUB-200-2011~\cite{cub200} (Section~\ref{sec:ablation}) are provided as empirical justifications. Next, we investigate how our embeddings are affected by class subsampling in Section~\ref{sec:sampling_ratio}, addressing the key scalability concern of softmax loss where training complexity is linear with the number of classes. Finally in Section~\ref{sec:exp_soa}, we show that our method outperforms state-of-the-art methods on several retrieval datasets.

\subsection{Implementation}

All experiments were executed using PyTorch and a Tesla V100 graphic card. We compare our method with common architectures used in metric learning including GoogleNet~\cite{googlenet2014}, GoogleNet with Batch Normalization~\cite{batchnorm}, and ResNet50~\cite{kaiming16}. We initialize our networks with pre-trained ImageNet ILSVRC-2012~\cite{imagenet_cvpr09} weights. We add a randomly initialized fully connected layer to the pool5 features of each architecture to learn embeddings of varying dimensionality. To simplify the sensitivity to the initialization of the fully connected layer, we add a Layer Normalization~\cite{layernorm} without additional parameters between the pool5 and fully connected layer (See Section~\ref{sec:layer_norm_exp} for the ablation study). We L2 normalize the embedding and class weights before Softmax and use a temperature of 0.05 consistently.

Unless otherwise stated, we first train for 1 epoch, updating only new parameters for better initialization. We then optimize all parameters for 30 epochs with a batch size of 75 with learning rate 0.01 with exception to CUB200 where we use learning rate of 0.001 due to overfitting. We construct our batch by sampling 25 examples per class for Cars196 and CUB200 and 5 examples per class for SOP and InShop (as only $\approx$ 5 images per class in dataset). We use SGD with momentum of 0.9, weight decay of 1e-4, and gamma of 0.1 to reduce learning rate at epoch 15. During training, we apply horizontal mirroring and random crops from 256x256 images; during testing we center crop from the 256x256 image. Following~\cite{angular}~\cite{nofusslearning}, we crop to 227x227 for GoogleNet, otherwise 224x224. Complete implementation details can be found in our source code repository.\footnote{Code: https://github.com/azgo14/classification\_metric\_learning.git}

\subsection{Loss Function Comparisons}

We compare our normalized softmax loss against existing metric learning losses. To focus on contributions from the loss functions, we leave comparisons against methods that ensemble models~\cite{yuan2016HDC, Xuan2018DeepRE}, modify the feature extractor architecture~\cite{attention_metric}, or propose complex activation paths between the featurizer and final embedding~\cite{abier} for Section~\ref{sec:exp_soa}.

We present Recall@K and NMI results on three standard retrieval datasets in Table~\ref{tab:googlenet_loss}, Table~\ref{tab:bninception_loss}, and Table~\ref{tab:resnet50_loss}, comparing against reported performance of methods trained with model architectures of GoogleNet, GoogleNet with Batch Normalization (BNInception), and ResNet50 respectively. For GoogleNet with Stanford Online Products only, we saw around a 1\% Recall@1 improvement by training all parameters from start with 10x learning rate on new parameters when compared with models trained with our standard finetuning procedure.

As shown, our approach compares very strongly against existing baselines. When fixing dimensionality to 512, we see that the performance improvements of our softmax embeddings across architectures mirror classification performance on ImageNet ILSVRC-2012. We hope our results help disentangle performance improvements of existing metric learning methods due to advancements in methodology versus changes of base feature models.

\begin{table*}
\begin{center}
\resizebox{0.9\linewidth}{!}{
\begin{tabular}{l | c c c c | c c c c c | c c  c c c}
\hline
 &  \multicolumn{4}{c}{SOP} &  \multicolumn{5}{c}{CARS-196} &  \multicolumn{5}{c}{CUB-200} \\
\hline
Recall@K &  1 & 10 & 100  & NMI & 1 & 2 & 4 & 8 & NMI & 1 & 2 &4 & 8 & NMI \\
\hline\hline
Contras.$^{128}$~\cite{songCVPR16}& 42.0 & 58.2 & 73.8 & - & 21.7 & 32.3 & 46.1 & 58.9 &-  &  26.4 & 37.7 & 49.8 & 62.3 & - \\
Lift. Struc$^{128}$~\cite{songCVPR16}& - & - & - & - & 49.0 & 60.3 & 72.1 & 81.5 & 55.0 &  47.2 & 58.9 & 70.2 & 80.2 & 55.6 \\
Lift. Struc$^{512}$~\cite{songCVPR16}& 62.1 & 79.8 & 91.3 & - & - & - & - & -  & - & - & - & - & - & - \\
Hist Loss$^{512}$~\cite{UstinovaNIPS16}  &  63.9 & 81.7 & 92.2 & - & - & - & - & - & - &  50.3 & 61.9 & 72.6 & 82.4 & - \\
Bin. Dev$^{512}$~\cite{UstinovaNIPS16}  &  65.5 & 82.3 & 92.3 & - & - & - & - & -  & - & 52.8 & 64.4 & 74.7 & \underline{83.9} & - \\
Npairs$^{512}$~\cite{npairNIPS2016}  &  67.7 & 83.8 & 93.0 & 88.1 &  - & - &- &- & - & - &- &- & -  & -  \\
Npairs$^{64}$~\cite{npairNIPS2016}  &  - & - & - & - & 71.1 & 79.7 & 86.5 & 91.6 & \underline{64.0} & 51.0 & 63.3 & 74.3 & 83.2 & 60.4  \\
Angular$^{512}$~\cite{angular} &  \textbf{70.9} & \textbf{85.0} & \textbf{93.5} & \textbf{88.6} & \underline{71.4} & \underline{81.4} & \underline{87.5} & \underline{92.1} & 63.2 &  \underline{54.7} & \underline{66.3} & \underline{76.0} & \underline{83.9} & \underline{61.1}  \\
\hline
NormSoftmax$^{512}$ & \underline{69.0} & \underline{84.5} & \underline{93.1} & \underline{88.2} & \textbf{75.2} & \textbf{84.7} & \textbf{90.4} & \textbf{94.2} & \textbf{64.5} & \textbf{55.3} & \textbf{67.0} & \textbf{77.6} & \textbf{85.4} & \textbf{62.8} \\

\end{tabular}
}
\caption{Recall@K and NMI across standard retrieval tasks. All methods are trained using GoogleNet for a fair comparison.}
\label{tab:googlenet_loss}
\bigskip

\resizebox{0.9\linewidth}{!}{

\begin{tabular}{l | c c c c  | c c c c c | c c c c c }
\hline
 &  \multicolumn{4}{c}{SOP} &  \multicolumn{5}{c}{CARS-196} &  \multicolumn{5}{c}{CUB-200} \\
\hline
Recall@K &  1 & 10 & 100  & NMI & 1 & 2 & 4 & 8 & NMI & 1 & 2 &4 & 8 & NMI \\
\hline\hline
Clustering$^{64}$~\cite{songCVPR17}  & 67.0 & 83.7& 93.2 & 89.5 & 58.1 & 70.6 & 80.3 & 87.8 &  59.0 & 48.2 & 61.4 & 71.8 & 81.9 & 59.2 \\
Proxy NCA$^{64}$\cite{nofusslearning}  &  73.7 & - & - & \textbf{90.6} & 73.2 & 82.4 & 86.4 & 88.7 & \underline{64.9} & 49.2 & 61.9 & 67.9 & 72.4 & \underline{59.5} \\
HTL$^{512}$~\cite{htl}  &  \textbf{74.8} & \textbf{88.3} & \underline{94.8} & - & \underline{81.4} & \underline{88.0} & \underline{92.7} & \underline{95.7} & - & \underline{57.1} & \underline{68.8} & \underline{78.7} & \underline{86.5} & - \\ 
\hline
NormSoftmax$^{512}$  & \underline{73.8} & \underline{88.1} & \textbf{95.0} & \underline{89.8} & \textbf{81.7} & \textbf{88.9} & \textbf{93.4} & \textbf{96.0} & \textbf{70.5} & \textbf{59.6} & \textbf{72.0} & \textbf{81.2} & \textbf{88.4} & \textbf{66.2} \\

\end{tabular}
}
\caption{Recall@K and NMI across standard retrieval tasks. All methods are trained using BNInception for a fair comparison}
\label{tab:bninception_loss}
\bigskip

\resizebox{0.9\linewidth}{!}{

\begin{tabular}{l | c c c c | c c c c c | c c c c c }
\hline
 &  \multicolumn{4}{c}{SOP} &  \multicolumn{5}{c}{CARS-196} &  \multicolumn{5}{c}{CUB-200} \\
\hline
Recall@K &  1 & 10 & 100 & NMI & 1 & 2 & 4 & 8 & NMI & 1 & 2 &4 & 8 & NMI \\
\hline\hline
Margin$^{128}$~\cite{samplingmatters} & 72.7 & 86.2 & 93.8 & \underline{90.7} & 79.6 & 86.5 & 91.9 & 95.1 &  69.1  &  \textbf{63.6} & \textbf{74.4} & \underline{83.1} & \textbf{90.0} & \underline{69.0} \\ 
\hline
NormSoftmax$^{128}$  & \underline{75.2} & \underline{88.7} & \underline{95.2} & 90.4 & \underline{81.6} & \underline{88.7} & \underline{93.4} & \underline{96.3} & \underline{72.9} &  56.5 & 69.6 & 79.9 & 87.6 & 66.9 \\
NormSoftmax$^{512}$  & \textbf{78.2} & \textbf{90.6} & \textbf{96.2}  & \textbf{91.0}  & \textbf{84.2} & \textbf{90.4} & \textbf{94.4} & \textbf{96.9} & \textbf{74.0} & \underline{61.3} & \underline{73.9} & \textbf{83.5} & \textbf{90.0} & \textbf{69.7} \\
\end{tabular}
}
\caption{Recall@K and NMI across standard retrieval tasks. All methods are trained using ResNet50 for a fair comparison}
\label{tab:resnet50_loss}
\end{center}
\end{table*}
\subsection{Embedding Dimensionality}
\label{sec:exp_dimension}

%

%
\begin{figure*}
\centering
\begin{minipage}{.45\textwidth}
  \centering
  \includegraphics[width=0.95\linewidth]{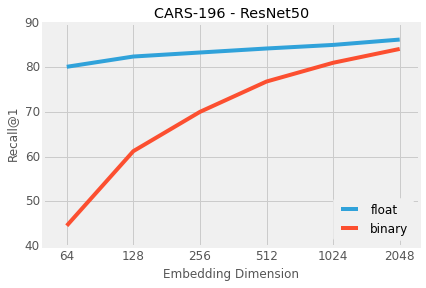}
  \end{minipage}%
  \begin{minipage}{.45\textwidth}
  \centering
  \includegraphics[width=0.95\linewidth]{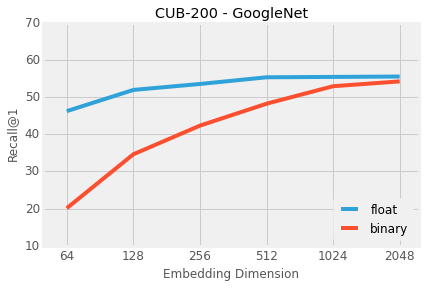}
  \end{minipage}
   \caption{Recall@1 for CARS-196 (left) and CUB-200-2011 (right) across varying embedding dimensions. Softmax based embeddings improve performance when increasing dimensionality. The performance gap between float and binary embeddings converge when increasing dimensionality, showing that when given enough representational freedom, Softmax learns bit like features.}
   \label{fig:dim_trend}

\end{figure*}

We study the effects of dimensionality on our classification-based embeddings by varying only the embedding dimension while keeping all other optimization parameters fixed. We have consistently observed that dimensionality is directly related to retrieval performance. Two examples of this across different datasets (CARS-196 and CUB-200-2011) and model architectures (ResNet50 and GoogleNet) are shown in Figure~\ref{fig:dim_trend}. Interestingly, this contradicts to reported behaviors for previous non-parametric metric learning methods~\cite{angular,songCVPR16,songCVPR17}, showing that dimensionality does not significantly affect retrieval performance. This difference is seen clearly when comparing R@1 across dimensionality for CUB-200-2011 with GoogleNet in Figure~\ref{fig:dim_trend} with the same dataset and model combination from ~\cite{songCVPR16}.

Higher dimensional embeddings lead to an increase in retrieval performance. Lower dimensional embeddings however are preferred for scalability to reduce storage and distance computation costs especially in large scale applications such as visual search~\cite{jing15}. We observe however that as we increase dimensionality of our embeddings, the optimizer does not fully utilize the higher dimensional metric space. Instead, we see that feature dimensions start relying less on the magnitude of each feature dimension and instead rely on the sign value. In Figure~\ref{fig:dim_trend}, we see for both datasets that the Recall@1 performance of binary features (thresholding the feature value at zero) converges with the performance of the float embeddings. This is a consistent result we see across datasets and model architectures. We show that training high dimensional embeddings and binarizing leads to the best trade-off of performance and scalability as described in Section~\ref{sec:exp_soa}.
\subsection{Ablation Studies}
\label{sec:ablation}

In this set of experiments we report the effect on Recall@1 with different design choices in our approach on CUB-200-2011. We train the ResNet50 with embedding dimension of 512 variant as in Table~\ref{tab:resnet50_loss}. We fix all hyperparameters besides the one of interest.

\begin{figure}
\centering
  \includegraphics[width=0.7\linewidth]{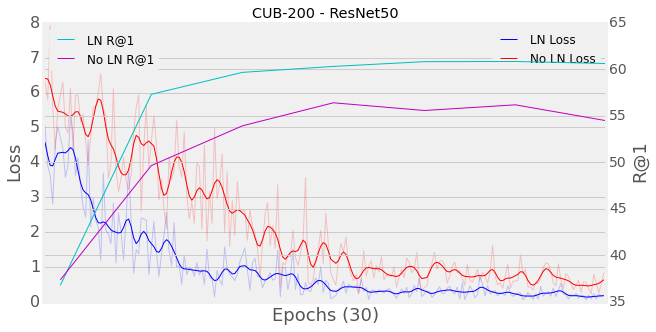}
   \caption{Loss and R@1 trends for training CUB-200 ResNet50 with and without Layer Normalization. Layer Normalization helps initialize learning, leading to better training convergence and R@1 performance.}
   \label{sec:label_norm_exp}

\end{figure}

\subsubsection{Layer Normalization}
\label{sec:layer_norm_exp}
We utilize Layer Normalization~\cite{layernorm} without parameters to standardize activation values after pooling to help the initialization of our training. With 100 classes in CUB-200-2011, we expect that a random classifier would have a loss value of roughly $-\ln(\frac{1}{100}) = 4.6$. As shown in Table~\ref{sec:label_norm_exp}, this occurs when training with Layer Normalization, but not without. We have found incorporating Layer Normalization in our training allows us to be robust against poor weight initialization of new parameters across model architectures, alleviating the need for careful weight initialization.

\begin{table}
\begin{center}
\resizebox{0.5\linewidth}{!}{
\begin{tabular}{l | c c c c c c c c}
\hline
S & -    & 1    & 3      & 12   & 25  & 37   & 75  \\
\hline
C & -    & 75   & 25     & 6   & 3    & 2    & 1 \\
\hline\hline
R@1 & 59.5 & 59.6 & 60.0  & 60.8 & 61.3 & 59.6 & 40.9 \\
\end{tabular}
}
\end{center}
\caption{ResNet50 Recall@1 on CUB-200-2011 dataset across varying samples per class for batch size of 75. (\textbf{S}) Samples per class in batch. (\textbf{C}) Distinct classes in batch. First column shows no class balancing in batch}
\label{tab:class_balance}
\end{table}

\subsubsection{Class Balanced Sampling}
\label{sec:class_balance_exp}
 As seen in Table~\ref{tab:class_balance}, class balanced sampling is beneficial over random sequential iteration over the dataset. Generally we've observed that performance improves when we have more samples per class until too few distinct classes exist in the minibatch where the bias of separating few distinct classes may introduce too much noise to the optimization, resulting in lower performance. For SOP and InShop, we simply sample roughly how many images exist per class on average ($\approx$ 5).



 \begin{figure}
 \centering
 \includegraphics[width=0.9\linewidth]{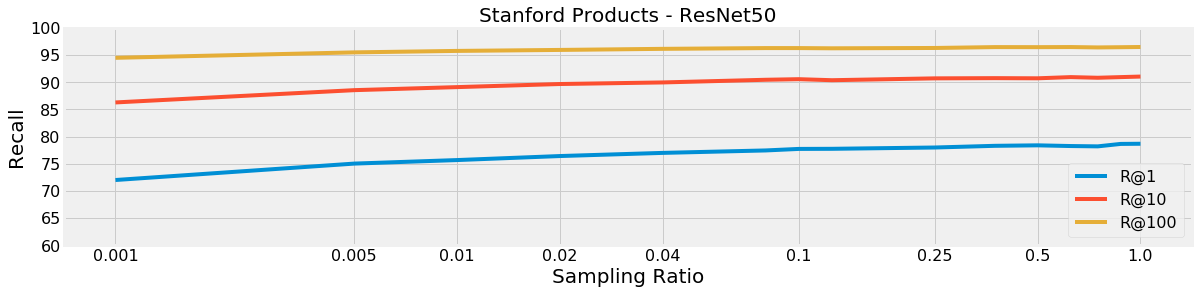}
 \caption{Recall@K for SOP with ResNet50 across class sampling ratios. We see that with sampling only 10\% of classes per iteration ($\sim$1K classes), we converge to a R@1 that is less than 1\% absolute drop in performance from using all classes.}
  \label{fig:sampling_ratio}
 \end{figure}

\subsection{Subsampling for Classification Scalability}
\label{sec:sampling_ratio}

To show that classification-based training is scalable, we apply the subsampling methodology to the Stanford Online Products dataset, which has the largest number of classes among datasets we tested. By subsampling, for each training minibatch the network only need to classify among a subset of all classes (randomly sampled subset which includes all classes within the training minibatch). We present our findings in Figure~\ref{fig:sampling_ratio}, showing that with only 10\% of the classes available during the forward pass of training, we can reach a R@1 performance comparable to using all classes (1\% drop in performance). When using 1\% of classes, we reach a R@1 of 75.7. When using 0.1\% of classes, we reach a R@1 of 72.0. As we can see, subsampling classes during training is an effective method of scaling softmax embedding training with little drop in performance.


\begin{table*}

\begin{center}
\resizebox{0.8\linewidth}{!}{
\begin{tabular}{l | c | c c c | c c c c }
\hline
 & Net. & \multicolumn{3}{c}{SOP} & \multicolumn{4}{c}{In-Shop}   \\
\hline
Recall@K &  & 1 & 10 & 100 & 1 & 10 & 20 & 30 \\
\hline\hline
Contrastive$^{128}$~\cite{songCVPR16}& G & 42.0 & 58.2 & 73.8 & - & - & - & - \\
Lifted Struct$^{512}$~\cite{songCVPR16}& G & 62.1 & 79.8 & 91.3 & - & - & - & - \\
Clustering$^{64}$~\cite{songCVPR17} & B & 67.0 & 83.7& - & - & - & - & - \\
Npairs$^{512}$~\cite{npairNIPS2016} & G & 67.7 & 83.8 & 93.0 & - & - & - & - \\
HDC$^{384}$~\cite{yuan2016HDC} & G & 69.5 & 84.4 & 92.8 & 62.1 & 84.9 & 89.0 & 91.2 \\
Angular Loss$^{512}$~\cite{angular} & G & 70.9 & 85.0 & 93.5 & - & - & - & - \\
Margin$^{128}$~\cite{samplingmatters} & R50 & 72.7 & 86.2 & 93.8 & - & - & - & - \\
Proxy NCA$^{64}$\cite{nofusslearning} & B & 73.7 & - & - & - & - & - & - \\ 
A-BIER$^{512}$~\cite{abier} & G  &  74.2 & 86.9 & 94.0 & 83.1 & 95.1 & 96.9 & 97.5 \\
HTL$^{512}$~\cite{htl} & B & 74.8 & 88.3 & 94.8 & - & - & - & - \\ 
HTL$^{128}$~\cite{htl} & B & - & - & - & - & 80.9 & 94.3 & 95.8 \\ 
ABE-8$^{512}$~\cite{attention_metric} & G$\dagger$ & 76.3 & 88.4 & 94.8 & 87.3 & 96.7 & 97.9 & 98.2 \\
DREML$^{9216}$~\cite{Xuan2018DeepRE} & R18 & - & - & - & 78.4 & 93.7 & 95.8 & 96.7 \\
LMCL$^{512}$~\cite{Wang2018CosFaceLM} & R50 & 60.9 & 76.5 & 87.0 & 73.3 & 90.5 & 93.1 & 94.3 \\
\hline
LMCL$^{\ast2048}$~\cite{Wang2018CosFaceLM} & R50 & 77.4 & 89.7 & 95.3 & \textbf{89.8} & \textbf{97.8} & \underline{98.6} & \underline{98.8} \\
NormSoftmax$^{1024}$ & B & 74.7 & 88.3 & 95.2 & 86.6 & 97.0 & 98.0 & 98.5 \\
NormSoftmax$^{128}$  & R50 & 75.2 & 88.7 & 95.2 & 86.6 & 96.8 & 97.8 & 98.3 \\
NormSoftmax$^{512}$  & R50 & \underline{78.2} & 90.6 & 96.2 & 88.6 & 97.5 & 98.4 & \underline{98.8} \\
NormSoftmax$^{2048}$ & R50  & \textbf{79.5} & \textbf{91.5} & \textbf{96.7} & \underline{89.4} & \textbf{97.8} & \textbf{98.7} & \textbf{99.0} \\
NormSoftmax$^{2048\textbf{bits}}$ & R50  & \underline{78.2} & \underline{90.9} & \underline{96.4} & 88.8 & \textbf{97.8} & 98.5 & \underline{98.8} \\
\hline
\end{tabular}
}
\end{center}
\caption{Recall@K on Stanford Online Products (SOP) and In-Shop. R - ResNet, G - GoogleNet, B - BNInception, $\dagger$ refers to refers to additional attention parameters, LMCL$^\ast$ is our method trained with the Loss}
\label{tab:metric_stanford}

\end{table*}

\begin{table*}
\begin{center}
\resizebox{0.8\linewidth}{!}{
\begin{tabular}{l | c | c c c c | c c c c }
\hline
& Net. & \multicolumn{4}{c}{CARS-196} & \multicolumn{4}{c}{CUB-200}   \\
\hline
Recall@K &  & 1 & 2 & 4 & 8 & 1 & 2 & 4 & 8 \\
\hline\hline
Contrastive$^{128}$~\cite{songCVPR16}& G  & 21.7 & 32.3 & 46.1 & 58.9 &  26.4 & 37.7 & 49.8 & 62.3 \\
Lifted Struct$^{128}$~\cite{songCVPR16}& G  & 49.0 & 60.3 & 72.1 & 81.5 & 47.2 & 58.9 & 70.2 & 80.2 \\
Clustering$^{64}$~\cite{songCVPR17} & B &  58.1 & 70.6 & 80.3 & 87.8 & 48.2 & 61.4 & 71.8 & 81.9 \\
Npairs$^{64}$~\cite{npairNIPS2016} & G  & 71.1 & 79.7 & 86.5 & 91.6 & 51.0 & 63.3 & 74.3 & 83.2 \\
Angular Loss$^{512}$~\cite{angular} & G  & 71.4 & 81.4 & 87.5 & 92.1 & 54.7 & 66.3 & 76.0 & 83.9 \\
Proxy NCA$^{64}$~\cite{nofusslearning} & B  & 73.2 & 82.4 & 86.4 & 88.7 & 49.2 & 61.9 & 67.9 & 72.4 \\
HDC$^{384}$~\cite{yuan2016HDC} & G  & 73.7 & 83.2 & 89.5 & 93.8 & 53.6 & 65.7 & 77.0 & 85.6  \\ 
Margin$^{128}$~\cite{samplingmatters} & R50  & 79.6 & 86.5 & 91.9 & 95.1 &  \underline{63.6} & 74.4 & 83.1 & 90.0 \\ 
HTL$^{512}$~\cite{htl} & B  & 81.4 & 88.0 & 92.7 & 95.7 & 57.1 & 68.8 & 78.7 & 86.5 \\ 
A-BIER$^{512}$~\cite{abier} & G &  82.0 & 89.0 & 93.2 & 96.1 & 57.5 & 68.7 & 78.3 & 86.2 \\
ABE-8$^{512}$~\cite{attention_metric} & G$\dagger$ & 85.2 & 90.5 & 94.0 & 96.1  & 60.6 & 71.5 & 79.8 & 87.4 \\
DREML$^{576}$~\cite{Xuan2018DeepRE} & R18 & 86.0 & 91.7 & 95.0 & 97.2 & 63.9 & 75.0 & 83.1 & 89.7 \\
LMCL$^{512}$~\cite{Wang2018CosFaceLM} & R50 & 73.9 & 81.7 & 87.4 & 91.5 & 58.7 & 70.3 & 79.9 & 86.9 \\
\hline
LMCL$^{\ast2048}$~\cite{Wang2018CosFaceLM} & R50 & 88.3 & 93.1 & 95.7 & 97.4 & 61.2 & 71.4 & 80.4 & 87.4 \\
NormSoftMax$^{1024}$ & B & 87.9 & 93.2 & 96.2 & 98.1 & 62.2 & 73.9 & 82.7 & 89.4 \\
NormSoftmax$^{128}$  & R50 & 81.6 & 88.7 & 93.4 & 96.3  & 56.5 & 69.6 & 79.9 & 87.6 \\
NormSoftmax$^{512}$  & R50 & 84.2 & 90.4 & 94.4 & 96.9  & 61.3 & 73.9 & 83.5 & 90.0 \\
NormSoftmax$^{2048}$ & R50 & \textbf{89.3} & \textbf{94.1} & \textbf{96.4} & \textbf{98.0} & \textbf{65.3} & \textbf{76.7} & \textbf{85.4} & \textbf{91.8} \\
NormSoftmax$^{2048\textbf{bits}}$ & R50 & \underline{88.7} & \underline{93.7} & \textbf{96.4} & \textbf{98.0} & 63.3 & \underline{75.2} & \underline{84.3} & \underline{91.0} \\ 
\hline
\end{tabular}
}
\end{center}
\caption{Recall@K on CARS-196 and CUB-200-2011. R - ResNet, G - GoogleNet, B - BNInception, $\dagger$ refers to additional attention parameters, LMCL$^\ast$ is our method trained with the Loss}
\label{tab:metric_cub200}
\end{table*}

\subsection{Comparison against State of the Art}
\label{sec:exp_soa}

Finally we compare our best performing softmax embedding model against state-of-the-art metric learning approaches on Stanford Online Products, In-Shop, CARS-196, and CUB-200-2011. We reproduced the state-of-the-art method in face verification literature, Large Maring Cosine Loss (LMCL)~\cite{Wang2018CosFaceLM}, and trained two variants of the networks: one with their recommended network architecture of 512 dimensional embeddings, and one with our modifications of 2048 dimensional embeddings. 


As shown in Table~\ref{tab:metric_stanford} and Table~\ref{tab:metric_cub200}, our 2048 dimensional ResNet50 embedding outperforms previous approaches. Considering the higher dimensionality of our embeddings, we also show that our 2048 \textbf{binary} embedding, sharing the same memory footprint of a 64 dimensional float embedding, similarly outperforms state-of-the-art baselines. These binary features were obtained by thresholding the float embedding features at zero as in Figure~\ref{fig:dim_trend}. Considering the scalability and performance of our binary softmax features along the simplicity of our approach, we believe softmax embeddings should be a strong baseline for future metric learning work.

\section{Conclusion}
In this paper, we show that classification-based metric learning approaches can achieve state-of-the-art not only in face verification but general image retrieval tasks. In the metric learning community, a diverse set of base networks for training embedding of different sizes are compared to one another. In our work, we conducted comparisons through extensive experimentation, and establish that normalized softmax loss is a strong baseline in a wide variety of settings. 
We investigate common critiques of classification training and high dimensionality for metric learning through subsampling, making our approach viable even for tasks with a very large number of classes and binarization, allowing us to achieve SOTA performance with  the  same  memory  footprint  as  64  dimensional float embeddings across the suite of retrieval tasks. We believe these practical benefits are valuable for large scale deep metric learning and real world applications.

\bibliography{egbib}
\end{document}